\documentclass[12pt]{article}

\usepackage[top=2.5cm, bottom=2.5cm, left=2.5cm, right=2.5cm]{geometry}
\usepackage{alltt}
\usepackage{graphicx}
\usepackage{color}
\usepackage{nicefrac}
\usepackage{subfigure}
\usepackage{amsmath}
\usepackage{amsthm}
\usepackage{amssymb}
\usepackage{float}
\usepackage{framed}
\usepackage{stmaryrd} 
\usepackage{tikz}
\usepackage{tkz-graph}
\usepackage{lmodern}
\usepackage{fancyvrb, relsize}
\usepackage{listings}
\usepackage{subfigure}
\usepackage{float}
\usepackage{multirow}
\usepackage{paralist}
\usepackage{alltt}
\usepackage{array}
\usepackage{xspace}

\usepackage{algorithm}
\usepackage[noend]{algpseudocode}
\makeatletter
\def\BState{\State\hskip-\ALG@thistlm}
\makeatother

\newcommand{\hide}[1]{}

\newcommand{\Manyopt}{\texttt{ManyOpt}\xspace}
\newcommand{\ZT}{\texttt{Z3}}
\newcommand{\ZTopt}{\texttt{$\nu$Z}}

\newcommand{\myparagraph}[1]{\smallskip\noindent {{\bf {#1}}\ }}

\newcommand\EatDot[1]{}

\begin{document}

\begin{center}
{\Large \bf \Manyopt: An Extensible Tool for Mixed, Non-Linear Optimization Through
SMT Solving}\\

\bigskip
\bigskip
{\small Andrea Callia D'Iddio and Michael Huth\\
Department of Computing, Imperial College London\\
London, SW7 2AZ, United Kingdom\\
$\{$a.callia-diddio14, m.huth$\}$@imperial.ac.uk}
\end{center}

\date{\today}

\bigskip
\begin{abstract}
Optimization of Mixed-Integer Non-Linear Programming (MINLP) supports 
important decisions in applications such as Chemical Process Engineering. But 
current solvers have limited ability for deductive reasoning or the use of 
domain-specific theories, and the management of integrality
constraints does not yet exploit automated reasoning tools such as SMT
solvers. This seems to limit both scalability and reach of such tools 
in practice. We therefore present a tool, \Manyopt, for MINLP optimization that enables 
experimentation with reduction techniques which transform a MINLP problem to 
feasibility checking realized by an SMT solver. \Manyopt is similar to the SAT 
solver ManySAT in that it runs a specified number of such reduction techniques 
in parallel to get the strongest result on a given MINLP problem. The tool is 
implemented in layers, which we may see as features and where reduction 
techniques are feature vectors. Some of these features are inspired by known 
MINLP techniques whereas others are novel and specific to SMT. Our experimental 
results on standard benchmarks demonstrate the benefits of this approach. The 
tool supports a variety of SMT solvers and is easily extensible with new 
features,  courtesy of its layered structure. For example, logical formulas for 
deductive reasoning are easily added to constrain further the optimization of a 
MINLP problem of interest. 
\end{abstract}

\section{Introduction}

MINLP problems are like the familiar linear programming models, except
that constraints or the objective function may contain non-linear
terms, and that some variables may not have real-valued type but an
integral or even binary one. The underlying decision problem is
undecidable in general. 
But solving MINLP problems is crucial in many areas of application, such as
Chemical Process Engineering. Current mathematical tools for solving
MINLP problems, for example ANTIGONE~\cite{antigone}, 
made important progress in solving problems that occur in
practice. But
they presently do not support \emph{deductive reasoning}
and \emph{incremental solving}~--~which could generate explanatory
scenarios and 
support ``what-if'' questions. They also don't yet exploit the
powerful reasoning techniques of SMT solvers in the transformation of
MINLP problems into problem types that offer computational
advantages~--~such as NLP whose problems have no integrality constraints.

State-of-the-art \emph{SMT solvers} do 
support deductive reasoning and incremental solving, but
they usually have little support for non-linear arithmetics and
arithmetics that combine real and integer types, as is the case in
MINLP problems. This motivated us to write a tool, called \Manyopt{},
whose purpose is to be an extensible workbench for exploring and
evaluating how MINLP problems can be reduced to instances of SMT
solving. The tool therefore inherits the ability of incremental
solving and deductive reasoning from SMT solvers. 

The default mode of our tool executes in parallel different reduction
techniques from MINLP to instances of SMT solving.
This parallelization means that these techniques compete on a given
input and so may speed up solving since the ``winner'' will terminate
optimization. Reduction techniques themselves are conceptualized as
feature vectors, and so are a composition of features that may
coordinate or complement different reduction activities. Our tool can
therefore be extended with new features and new feature vectors to get
richer parallelized solvers for MINLP problems.  The tool makes the
assumption that optimal values are only computed up to some specified
accuracy, a further input parameter to the tool. 

Reduction techniques need to deal with computational 
problems related to the management of integrality constraints and to
the non-linearity of terms. The tool \Manyopt{} deals with
such integrality management by using the parallel combination of 
dynamic branch and bound techniques and a novel static \emph{binarized 
flattening} technique. The experiments reported in this paper
demonstrate that the combination of such techniques is instrumental in
solving many MINLP benchmarks that would otherwise not be solvable
within reasonable running times by
branch and bound techniques that reduce to SMT problem instances.

The non-linearity of terms is dealt with 
by the parallel combination of a naive search technique, 
which has shown to often work well with satisfiability solving (see MiniSAT+ 
\cite{minisat}), with an unbounded binary search technique, which is faster in 
scanning intervals containing many feasible objective values. Our tool
also supports the combination of these
two techniques into a hybrid approach whose cooperative strategy
conjoins
the invariants of both techniques. Our experiments provide evidence
that the approach we advocate and develop here has definite potential
in crafting MINLP solvers based on SMT.

\label{section:introduction}

\paragraph{Contributions of our paper:} 
We offer a tool with which researchers can explore techniques for
reducing MINLP optimization problems to instances of SMT, where
optimization is computed up to some specifiable accuracy. We organize
the design space for such reduction techniques such that points in
that space are feature vectors, where features are choices within a
particular reduction layer of that tool. These layers are easily
extensible with new features and the tool supports, in principal, the
use of any SMTLIB 2.0 compliant SMT solver. The presently implemented
features of \Manyopt{} represent some techniques already familiar in the
optimization community but also novel techniques that are made
possible by using an SMT solver as a feasibility checker. We evaluate
this tool through standard MINLP benchmarks. In particular, we show
that our tool can solve some benchmarks from Chemical Process
Engineering that are widely held to be challenging for existing MINLP
solvers: our tool solves all these problems and with higher accuracy
than that reported in some existing MINLP tools. We also evaluate
\Manyopt{} on 193 benchmarks from the MINLPLIB library, 138 NLP
problems and 56 MINLP problems. Our tool solved 68\% of the MINLP
problems and 63\% of the NLP problems within 30 minutes.
This appears to be the strongest evidence yet of the
potential of SMT solving for MINLP problems. The analysis of these
experiments also clearly reveals the benefits of running different
reduction techniques in parallel. For example, 
using just one
reduction technique for dealing with non-linearity would dramatically
reduce the percentage of solved benchmarks and increase the computing
time for solving.

\paragraph{Outline of paper} In Section~\ref{section:background} we
recall some background on MINLP problem types and SMT. The tool
architecture and functionality of \Manyopt{} is the topic of
Section~\ref{sec:architecture}. A description of the algorithms and
supported features is given in Section~\ref{section:description}.
In Section~\ref{sec:experiments} we evaluate our approach and tool
experimentally. Related work is discussed in
Section~\ref{section:related} and the paper concludes in Section~\ref{section:conclusion}.

\section{Background}
\label{section:background}

\myparagraph{Mixed Integer Non-linear optimization.}
A \emph{Mixed Integer Non-Linear Programming} (\emph{MINLP}) problem is a mixed 
integer optimization problem where the objective function,  some constraints or 
both may be non-linear \cite{floudas1995nonlinear}. This leads to established 
subtypes of MINLP: In LP and NLP problems, there are no integrality constraints. 
In LP, the objective function and the constraints are linear, in NLP problems 
they can also be non-linear. ILP and INLP problems are variants of LP and NLP, 
respectively, in which all variables must have integer values. MILP is a subtype 
of MINLP in which constraints and objective function are linear. BLP, BNLP, MBLP 
and MBNLP correspond to ILP, INLP, MILP, and MINLP, respectively, but 
integrality constraints are replaced by ``binarity'' constraints: all 
variables from the designated set must have binary instead of integral 
values.

\myparagraph{Complexity analysis.}
Unfortunately, the problem of solving a generic MINLP problem to deterministic 
global optimality is \emph{undecidable} \cite{MINLPDecidability}, due to a 
combination of non-linearity and integrality. Notwithstanding, several efforts 
have been made to solve MINLP problems, supported by decomposition or 
transformation techniques specific to subclasses of MINLP: Solving a generic NLP 
problem has been proven to be NP-hard \cite{NLPComplexity}, but many effective 
methods and algorithms have been developed in the last four decades (see 
\cite{bazaraa2013nonlinear} for a survey). Solving a generic ILP or MILP problem 
is also NP-hard \cite{MILP,MILPComplexity}, but again much progress has been 
done to develop effective techniques (see \cite{MILP} for example). 

\myparagraph{Optimization based on SMT.}
Satisfiability Modulo Theories (SMT) solvers \cite{SMT} are an
extension of SAT solvers,
where additional \emph{theories} can be used to express logical formulae and 
to find a satisfying assignment. The SMT solvers contain specific algorithms
based on many different theories (see e.g.\ \cite{DecisionProcedures}), including:
theory of equality,
bitvectors,
reals,
integers,
and arrays.
An important and popular SMT solver among the ones supported directly in our 
tool is \ZT\  \cite{Z3}. \ZT\  is an efficient \emph{incremental} SMT solver 
that supports \emph{exact arithmetic}  \cite{exact}, which is fundamental 
whenever exact solutions or high confidence in approximate solutions is 
required.

The theories modeled in SMT and the related languages provide support to 
naturally encode the variables and the constraints of an optimization problem, 
thus SMT solvers can be used as a black box in order to act as a feasibility 
solver for optimization algorithms. 

In \cite{bjorner2014nuz} the \ZT\  solver has been extended in order to also
perform optimization. The resulting solver, called \ZTopt, can be used to
combine optimization with the powerful capabilities of \ZT.
The \ZTopt\  solver has shown the ability to efficiently solve many MILP 
optimization problems and it competes well with other SMT-based optimizers 
presented in \cite{sebastiani2015optimization} and \cite{li2014symbolic} in this 
area.
But \ZTopt\  supports only linear mixed arithmetic, thus it has very little 
support for MINLP problems as such. The main purpose of our research is to add 
such capability in the context of SMT-based optimization.

\section{Tool Architecture and Functionality}
\label{sec:architecture}

\begin{figure}
\begin{center}
\includegraphics[scale=0.7]{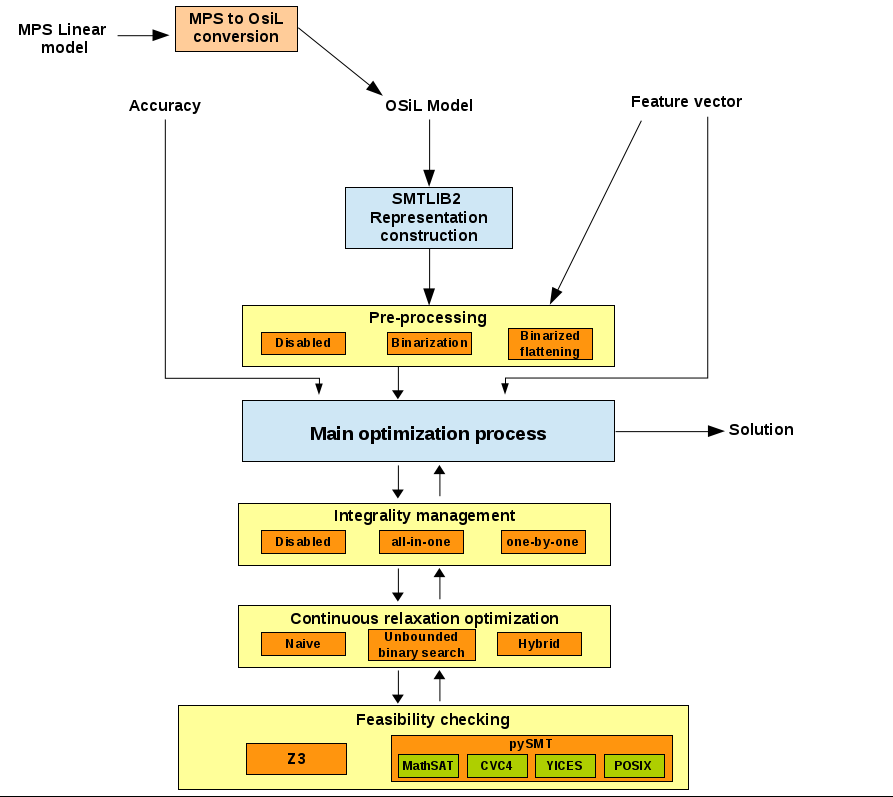}
\end{center}
\caption{The architecture and approach of our tool \Manyopt{}
\label{fig:structure}}
\end{figure}

Figure \ref{fig:structure} shows the main structure of our tool \Manyopt, 
represented as a data flow diagram. As input it takes a desired \emph{accuracy} 
for the computed global optimum and, either an OSiL model for a MINLP problem or an MPS model for a MILP 
problem which is translated into OSiL. Then \Manyopt executes \emph{in 
parallel}  a set of dataflows on that OSiL input, where each dataflow has its own 
feature vector of how to realize configurable tool layers. The tool layers are 
Preprocessing, Integrality Management, Continuous Relaxation Optimization, and 
Feasibility Checking. Each of these layers contains within it an (extensible) 
list of features, as seen in Fig.~\ref{fig:structure}. The default setting uses 
18 feature vectors run as parallel processes and keeps track of dependencies 
within each such vector. For example, if feature BinarizedFlattening is enabled 
in Preprocessing, then feature Disabled has to be chosen for Integrality 
Management.

For each feature vector, the data flow of its process is as follows. First, an 
SMTLIB~2.0 representation of the model is constructed, based on the input and 
selected features. After that, a \emph{Preprocessing} phase may transform the 
SMTLIB~2.0 representation using either \emph{Binarization} or our novel 
\emph{Binarized Flattening} techniques to convert the MINLP problem into a MBNLP 
problem, where all integer variables can only have values 0 or 1. 

Second, the Main Optimization Process uses the lower three layers, organized as 
a \emph{stack}, to find an optimal solution for the given model within the 
specified accuracy. The \emph{Integrality Management} layer may apply 
\emph{branch-and-bound} techniques such as One-By-One or our novel All-In-One to 
handle integrality constraints. This layer produces NLP problems as continuous 
relaxations of the original MINLP problem, and then relies on the 
\emph{Continuous Relaxation Optimization} layer to solve them. Alternatively, 
Integrality Management may be disabled, for example if the original problem is 
already an NLP problem, or if we wish to move the complexity of integrality 
management to lower layers.

The Continuous Relaxation Optimization layer takes an NLP problem as input and 
reduces optimization to the repeated use of the \emph{Feasibility Checking} 
layer. A number of such reduction algorithms are supported: the \emph{naive} 
method (the term ``naive'' is used without prejudice and inspired by work on 
pseudo-boolean optimization \cite{minisat}), the \emph{unbounded binary search} 
method, and a novel method of us based on a \emph{hybrid} combination of these 
two methods.

The feasibility of an NLP problem, represented in SMTLIB~2.0 in our tool, is 
verified using an SMT solver. Currently, \Manyopt supports 
in the feasibility layer in Figure \ref{fig:structure}, the solver \texttt{Z3} 
\cite{Z3} directly, the solvers \texttt{MathSAT} \cite{MathSAT}, \texttt{CVC4} 
\cite{CVC4} and \texttt{YICES} \cite{YICES} through the \texttt{PySMT} library 
\cite{PySMT}; other SMT solvers compatible with SMTLIB~2.0 may be used 
through a \emph{POSIX piping}.

\section{Tool description: algorithms and software design}
\label{section:description}

\subsection{Input parsing: OSiL and MPS}

Our \Manyopt tool supports two input formats for optimization models. The OSiL 
format \cite{OSiL} has been defined and published by COIN-OR within their 
Optimization Services project \cite{OS}. It is a widely adopted XML representation
(e.g.\ in \cite{MINLPLIB}) of an optimization problem, and its schema 
is available in \cite{OSiLSchema}. 

The OSiL language distinguishes among linear constraints, quadratic constraints 
and generic non-linear constraints. Linear constraints are represented using a 
compact coefficient matrix, and quadratic constraints are represented by a tuple 
list. Non-linear constraints are represented by using trees, where each node 
represents an operator and its children represent the operands. 

Using an XML representation for MINLP problems has several benefits in
terms of robustness, stability and standardizability. In particular, using an
XML schema facilitates checking for errors in an OSiL model without the need
to write code for this purpose.
Given an OSiL model as input, \Manyopt constructs several alternative SMTLIB 2.0
representations of it, whose construction is based on the configuration values
in each feature vector corresponding to a specific parallel execution.

Our tool also supports Mixed Integer \emph{Linear} Programming models (MILP)
represented in MPS format, which is another well-known format for optimization
(see e.g.\ \cite{MIPLIB}), still consisting of a compact coefficient matrix
representing linear constraints. We added this support by writing a simple 
\texttt{C++} code entirely based on the conversion libraries provided by
COIN-OR, in order to convert the MPS input into OSiL format.

Our tool OSiL parser can detect whether the problem is MINLP or just NLP, and in 
the second case it can enable in \Manyopt{} only those feature vectors that are 
useful for NLP, thus saving system resources and improving performance.

\subsection{Pre-processing: binarization and flattening}
\myparagraph{Binarization} Binarization mechanisms are widely used for mixed 
optimization, in order to handle integrality constraints where variables 
can only have values 0 or 1 (see \cite{floudas1995nonlinear} Subsection 6.2.1), 
thus transforming the MINLP problem into an MBNLP problem. In order to achieve 
this, each variable involved in an integrality constraint is represented by 
creating a set of variables representing the ``bits'' of the original variable,
which is assumed to be bounded in order to determine the amount of ``bits''
needed. Then $ 0 $ and $ 1 $ are set as lower and upper bounds for these
new variables, and they are constrained to be integer, i.e.\ either $ 0 $ or
$ 1 $.

Formally, an integrality constraint $ x \in \mathbb{Z} $ for variable $ x $ 
bounded by $l\leq x\leq u$ is represented by $q$ many variables $b_1,\dots, 
b_q$ with $q = 1 + \lceil \log_2 (u - l) \rceil$, equality 
constraint $x = l + b_1 + 2b_2 + 4b_3 + \dots + 2^{q-1}b_q$ and inequalities 
$0\leq b_i\leq 1$ for all $i$ such that $1\leq i\leq q$ added to the 
optimization problem. Now, the integrality constraints $ b_i \in \mathbb{Z} $ 
can replace the original $x\in \mathbb{Z}$ integrality constraint.

\paragraph{Binarized Flattening} An alternative to branch and bound consists in 
\emph{flattening} the integrality constraints during the pre-processing stage, 
and moving the complexity to the logic manipulation abilities provided by the 
SMT solvers by using the OR operator. More precisely, this method can be 
imagined as a sort of ``extreme'' version of branch and bound, which tries to 
prevent all the integrality violations by adding specific constraints before 
starting the actual solving of the optimization problem. After that, a simple 
optimization without any branch and bound method can be started, thus the 
Integrality Management layer can be set to ``Disabled''. This approach also
assumes that the variables involved in integrality constraints are bounded.

Formally, for each integrality constraint $ x \in \mathbb{Z} $ where
$ x $ is bounded by $ l\leq x\leq u $ we add the assertion
\begin{equation}
(x \geq l) \wedge \bigwedge_{l \leq i \leq u} (x \leq i \vee x \geq i
+ 1) \wedge (x \leq u)
\end{equation}
If this approach is applied as it is, then it would be dramatically inefficient. 
In particular, for each variable $ x $ involved in integrality constraints, it 
generates an assertion containing $ size(x) + 1 $ disjunctions where $size(x) $ 
is $ u - l $, where $ u $ and $ l $ still represent the upper and the lower 
bound for $ x $. This means that if the lower bound of $ x $ is $ -10000 $ and 
the upper bound is $ + 10000 $ (realistic numbers for many problems), this adds 
an assertion with $ 20001 $ disjunctions. For hundreds of such integrality
constraints, the size of the resulting new feasibility problem would be hard to
manage.

But, if this approach is combined with binarization, then each variable
involved in integrality constraints would be bounded just by 0 and 1, then
only one disjunction is needed to exclude all the non-integer values
between 0 and 1. In particular, if $ b \in \mathbb{Z} $ is an integrality
constraint in a binarized problem, then adding the assertion
\begin{equation}
(b = 0) \vee (b = 1)
\end{equation}
is sufficient to ``flatten'' it. One may wonder whether the fact that such a simple
assertion can be added to flatten a binarized program is counter-balanced by
the fact that many variables may be needed to represent the original variables
as sequences of bits.

For the above example, binarization needs 
only $ \lceil \log_2(20001) \rceil = 18 $ ``binary'' variables to represent it with $ size(x) = 20001 $.
Since just one disjunction is used to flatten each ``binary'' variable, only $ 18 $
disjunctions need to be added instead of $ 20001 $ if flattening is applied to the binary representation of $ x $.

To the best of our knowledge, binarized flattening is a novel approach and the 
experimental results in Section \ref{sec:experiments} clearly show that it has 
been the most valuable method implemented in the \Manyopt tool, as it has been
fundamental to efficiently solve many problems which have not been solved 
within given timeouts by the methods based on branch and bound.

\subsection{Integrality management: one-by-one vs all-in-one}
Most of the methods we implemented for integrality management are based on 
branch and bound techniques, but here we also move part of the combinatorial 
complexity to the lower layers.

\myparagraph{The one-by-one approach} This approach is similar to a standard 
branch and bound approach. But there is an important difference: assume that a
feasible solution is found in which at least one integrality constraint is
violated. Then choose one variable $ x $ whose value $ v $ violates one of
the integrality constraints and add the assertion $ x \leq \lfloor v 
\rfloor \vee x \geq \lceil v \rceil $ to the optimization problem. Since 
disjunctions are supported in SMT as assertions, this is well defined and avoids 
a split into two optimization problems by delegating such combinatorial 
complexity to the SAT engine of the SMT solver.

\myparagraph{The all-in-one approach} We call the above approach one-by-one, as 
each iteration only adds a sole assertion about a sole variable. But SMT solvers 
allow us to add more than one constraint at a time, leading to the 
\emph{all-in-one} approach; it collects \emph{all} variables whose values 
violate integrality constraints in a feasible solution and adds the above 
disjunction as  an assertion but for \emph{all} such variables 
simultaneously.

\myparagraph{Delegating to lower layers} As explained in section 
\ref{sec:architecture}, it is also possible to disable integrality management. 
In this case, the whole complexity due to integrality constraints is moved 
directly to lower levels. Considering the current components of the 
optimization stack in \Manyopt{}, this would mean that such complexity is moved to the 
Feasibility Checking layer, where SMT solvers can provide support since the 
variables involved in integrality constraints can be represented using 
\emph{integer variables} directly in SMT, and theories for mixed integer 
arithmetic are used by the SMT solver when it is called for feasibility 
checking.

\subsection{Continuous relaxation optimization}
As explained in section \ref{sec:architecture}, this layer provides algorithms 
to solve NLP problems, and it relies on the feasibility checking
layer.

\myparagraph{The naive approach}
This method is quite simple but it can be effective in many circumstances, and 
it has been used for optimization. It consists of a loop in which a value is 
found for the objective function using the feasibility checker, and an attempt 
to find a lower value is made by adding an assertion saying that the objective 
function must be smaller than that value. When the problem will become 
infeasible, then the last found objective value, if there is one, is the optimal 
solution. See \cite{minisat} for a more detailed description of 
this method.

\myparagraph{Unbounded binary search}
We use here unbounded binary search as developed and used 
in~\cite{beaumontconfidence} for optimization using an SMT solver. The unbounded 
binary search is composed of two main phases.
\begin{enumerate}
\item The \emph{bounds search}, in which initial lower and upper bounds are 
found such that the optimal value of the objective is between such bounds.
\item The \emph{bisection phase}, which is the actual binary search, where the 
interval between the lower bound and the upper bound is restricted by splitting 
it in two equal parts, until the optimal value of the objective is found, 
relative to the specified accuracy.
\end{enumerate}
For more details about this method, see \cite{beaumontconfidence}.

\myparagraph{Hybrid method}
Our novel Hybrid method exploits the fact that Naive and Unbounded Binary Search
methods can be written such that they have a similar structure and they share
the same invariants. It is still composed of two phases like unbounded binary
search, where first upper and lower bounds are determined for the objective,
then the interval is restricted to find the optimal value. But in each phase,
``naive steps'' are used to determine whether an interval is empty in order to
stop the algorithm if the current solution is already the optimal one.

Intuitively, the performance of this method on a specific NLP problem is 
expected to be closer to the performance of the best method (naive or 
unbounded binary search) on the same problem. This is confirmed by almost all the 
experiments in Section \ref{sec:experiments}.

\subsection{Feasibility checking: SMT solving}

Originally, the \Manyopt{} tool was relying exclusively on \ZT\ as a feasibility checker,
since \ZT\ provides a well-documented Python API which has been used by our
algorithms. But later, thanks to the PySMT framework \cite{PySMT}, we have
refactored the project and we made it independent of the underlying SMT solver,
by creating a Feasibility Checking layer in which several solvers are available.
More precisely, by using PySMT we now also support the SMT solvers MathSAT
\cite{MathSAT}, CVC4 \cite{CVC4} and YICES \cite{YICES}.

Among the above-mentioned solvers, \ZT\ is still the only one which has a
relevant support for non-linear arithmetics, especially the non-mixed one, i.e.\
with real arithmetics only. The other solvers provide a good support for linear
arithmetics and can be used for MILP problems. Besides, they implicitly provide
some support for MINLP problems where non-linearity is due to particular
combinations of linear functions, e.g.\ absolute values.

But, courtesy to the PySMT library, the \Manyopt tool also supports any
external SMTLIB 2.0 compliant solver through POSIX piping by just providing the
solver executable file. Therefore, \Manyopt can be used with SMT solvers
specifically designed to reason about non-linear real arithmetics, like
\cite{nonlinear3}, \cite{nonlinear2}, \cite{nonlinear1}.
Such solvers have not been tested yet, but it will be interesting to
use them in future work as feasibility checker with a parallel combination of our techniques
for Integrality Management and Continuous Relaxation Optimization.

\section{Experimental results and validation}
\label{sec:experiments}

Here we discuss the salient experimental results which have been obtained using the 
\Manyopt tool. The main source of benchmarks for our experiments is MINLPLIB 2.0 
\cite{MINLPLIB}, which contains a collection of more than one thousand MINLP and 
NLP problems in many formats, including the OSiL format we are able to parse in 
our tool. We also tested some MILP benchmarks available in MPS format from the 
MIPLIB2010 portal \cite{MIPLIB}, and we also ran the tool \Manyopt{} with some complex 
NLP benchmarks from the Chemical Engineering world. All the findings are 
reported below.

\subsection{Overall evaluation for \Manyopt}
We started the experimentation with \Manyopt by taking the benchmarks from
MINLPLIB and by starting to run all of them sorted by the benchmark
size (smallest first), in
order to be able to start with a reasonable timeout (30 minutes) for \Manyopt to
get the solutions. The experimentation is still in progress, and if needed we
will also increase the timeout as the benchmarks will increase in size and
complexity. In this case, the ``size'' of a benchmark has been chosen to be the
OSiL file size, in order to consider at the same time the number of variables,
the number of constraints and the complexity of constraints.

So far we have ran all the first 193 experiments in the MINLPLIB archive sorted 
by size, not including the ones which have been filtered out because they were 
containing transcendental functions. The exponential functions are the only
non-polynomial functions accepted by our tool, but usually they are not solved
because the SMT solving tools have almost no support for exponential functions.
But we included them in the experimentation.

The machine used for the experiments is a 12-core Intel\textregistered 
Xeon\textregistered E5-2640 2.50 GHz CPU with 47GB of RAM memory.
Of the 193 MINLPLIB experiments ran so far, 56 were MINLP and 138 were NLP.
For a fair analysis, we split up the experimental session into
two separate ones in order to avoid running one category more than the 
other one (e.g.\ NLP more than MINLP). All the complete logs from the experiments
are available in the \Manyopt tool webpage \cite{manyoptURL}, , where for each experiment many details are logged depending
on the chosen features, e.g. the bisection time for unbounded binary search, or
the number of branch and bound iterations.

For the 193 experiments we ran, excluding only one MINLPLIB 
benchmark,
the \Manyopt always returned the right solution when not exceeding the
timeout.
Clearly, the solution provided by 
\Manyopt was right with respect to the given accuracy, which has been set to $ 
0.001 $.

We ran our experiments using 18 feature vectors in parallel for MINLP
processes, by choosing all the possible methods in each layer, except for the
feasibility layer, where we just chose \ZT\ because it can support non-linear
arithmetic.

For the NLP problems, preprocessing and integrality management features are
not applicable because there are no integrality constraints. Thus, \Manyopt runs
only 3 feature vectors in parallel for those problems. 

As said above, the logs produced by the experiments are all available in 
the \Manyopt tool webpage \cite{manyoptURL}. Below we summarize the most 
important facts conveyed by the experimentation.

\begin{itemize}
\item Regarding MINLP problems, our tool found the right solution within the
given timeout, in about 68\% of the problems. 
\item Our tool solved about 63\% of the NLP problems within the given timeout. 
\end{itemize}

These are very high percentages, since the underlying \ZT\ solver used for the
experimentation has low support for non-linear and mixed arithmetic. Plus
the percentage is surely higher if we also filtered out some problems containing
exponential expressions; we will soon find a way to automatically filter them out
and the updated results will be shown on the tool website.

\myparagraph{Observation}
Compared to the current literature, this is an important step forward in
supporting non-linear and mixed constraints in SMT-based optimization. The 
complementarity of our methods was crucial to obtain this; we analyze these methods experimentally layer by layer as discussed below.

\paragraph{}
We also tried to solve some complex and widely believed to be
challenging
problems from Chemical Engineering, 
still with a 30 minutes timeout, and we solved 4 different problems in 
\cite{osil-castro,bental,haverly}. In particular, \cite{osil-castro} 
reports that a popular non-linear solver like MINOS \cite{minos} provided a poor 
solution for the example 2 in that paper, while our tool found the right 
solution within the timeout.

\newcommand{\notapplicable}{*}
\newcommand{\NA}{\qed}
\newcommand{\UN}{$\blacksquare$}

\subsection{Layer analysis: preprocessing and integrality management}

\begin{figure}
\centering
\begin{scriptsize}
\begin{tabular}{|l|r|r|r|r|r|r|r|}
\hline
\textbf{benchmark}     & \textbf{allinone} & \textbf{onebyone} & \textbf{nobb} & \textbf{bin\_allinone} & \textbf{bin\_onebyone} & \textbf{bin\_flattening} \\
\hline                 
hmittelman             &    \NA            &    \NA            &     \NA      &        \NA             &         \NA            &          1.08            \\
nvs04                  &     0.38          &     0.46          &     \UN      &        \NA             &         \NA            &          \NA             \\
nvs06                  &     0.81          &     0.39          &     \UN      &        \NA             &         \NA            &          1.08            \\
nvs07                  &     0.21          &     5.59          &     \UN      &        12.22           &         \NA            &          0.19            \\
nvs10                  &     0.23          &    \NA            &     \NA      &       149.79           &         \NA            &          0.06            \\
nvs16                  &     0.27          &     0.27          &     \UN      &        \NA             &         \NA            &          \NA             \\
gear                   &     0.28          &     0.33          &     \UN      &        2.11            &         4.45           &        112.07            \\
prob02                 &    \NA            &    \NA            &     \UN      &        \NA             &         \NA            &         15.54            \\
prob03                 &    \NA            &    \NA            &     \UN      &        \NA             &         \NA            &          0.02            \\
st\_miqp3              &    \NA            &    \NA            &      0.03    &        \NA             &         \NA            &          \NA             \\
nvs15                  &     2.85          &    \NA            &     \UN      &        \NA             &         \NA            &          0.03            \\
st\_testph4            &    \NA            &    \NA            &     \UN      &        \NA             &         \NA            &          0.06            \\
st\_miqp2              &     0.24          &     0.49          &     \UN      &        \NA             &         \NA            &         15.28            \\
st\_miqp1              &    \NA            &    \NA            &     \UN      &        \NA             &         \NA            &          0.05            \\
st\_test1              &     0.17          &    \NA            &     \UN      &         3.11           &         \NA            &          0.03            \\
st\_test2              &    \NA            &    \NA            &     \NA      &        \NA             &         \NA            &         16.38            \\
ex1221                 &     0.48          &     1.17          &     \NA      &         0.65           &          2.45          &          0.19            \\
fac1                   &    43.58          &   120.12          &     \UN      &        \NA             &         \NA            &         14.70            \\
st\_e13                &    14.09          &    14.11          &     \UN      &        22.12           &         41.21          &          0.02            \\
sporttournament06      &    \NA            &    \NA            &     \UN      &        \NA             &         \NA            &          4.75            \\
gbd                    &     0.03          &     0.03          &      0.02    &         0.04           &          0.04          &          0.02            \\
st\_e27                &    \NA            &    \NA            &     \NA      &        \NA             &         \NA            &          0.16            \\
alan                   &     0.11          &     0.23          &     \NA      &         0.14           &          0.55          &          0.09            \\
nvs01                  &     1.55          &     2.55          &     \UN      &        \NA             &         \NA            &          \NA             \\
gear4                  &    \NA            &    \NA            &     \UN      &        19.12           &         91.31          &         15.92            \\
tln2                   &    \NA            &    10.32          &     \UN      &         0.90           &         16.64          &          0.07            \\
ex1263a                &    \NA            &    \NA            &     \UN      &        \NA             &         \NA            &          1.16            \\
\hline                       
\end{tabular}
\end{scriptsize}
\caption{Some experimental data about preprocessing and integrality
  management methods: numbers show running times in seconds, 
white squares mean a timeout and in column {\bf nobb} black squares
refer to an ${\tt
    UNKNOWN}$ result of an SMT instance} \label{fig:integrality} \end{figure}

After analyzing the overall experimental results for \Manyopt, here we show some 
experimental data in which single feature vectors have been ran instead of the 
whole \Manyopt tool, in order to show the performances of each feature in the 
\Manyopt layers. Because of time constraints, since such single feature 
vectors have been executed one by one sequentially, for these experiments the 
timeout has been reduced to 150 seconds (otherwise, since 18 feature 
vectors have been tested, each single experiment could have taken up to 9 
hours).

For some representative benchmarks, Figure \ref{fig:integrality} 
compares the performance of all the methods from the Preprocessing and 
Integrality Management layers, since both layers manipulate integrality 
constraints. Each column represents a combination of methods for these two 
layers. More precisely, the columns show non-binarized all-in-one and one-by-one 
methods, the disabled integrality management (represented with {\bf nobb}, i.e.\ no 
branch and bound), which just delegates the integrality management to the SMT
solver, the binarized all-in-one and one-by-one methods and finally 
the binarized flattening method. For each combination, the best result obtained 
when combining it with the three methods for continuous relaxation optimization
is reported.

The little white squares in the paper represent a case in which the corresponding
combination of features did not manage to find the solution within the timeout.
In the {\bf nobb} column (i.e.\ disabled integrality management), a black square
means that the SMT solver returned ${\tt UNKNOWN}$ because of some incomplete theory
related to mixed and non-linear arithmetic.

The benchmarks reported here have been chosen to show behaviors which occur
in general in the experiments made so far, and the main observations
to make based on these results are the following:

The first thing to observe is the importance of the \emph{binarized flattening} 
method. The figure shows many cases (see e.g.\ hmittelman, prob02, prob03, 
st\_testph4, and others) in which binarized flattening has been the only method 
which managed to obtain the optimum efficiently, while all the other ones 
exceeded the timeout. Binarized flattening has been a crucial method in order to 
solve most of the problems even if \ZT\ has low support for non-linear and mixed 
arithmetic.

Looking at the table, we notice that the non-binarized 
\emph{all-in-one} approach also has an important role in solving. In many cases, 
it performs better than binarized flattening, or it solves the problem while 
binarized flattening exceeds the timeout. Together with binarized flattening, 
the non-binarized all-in-one mode is fundamental for finding the solution for an 
important fraction of the MINLP problems tested so far. The complementarity 
relationship between these two approaches is clear (one is binarized, the other 
one is non-binarized, one applies branch and bound dynamically, the other one 
statically flattens all the integrality conditions a priori, etc.), and their 
parallel execution lets us get the benefits from both of them, depending on the 
type of optimization problem which must be solved.

The other features shown in the table can also win occasionally. For example, 
the {\bf nobb} approach, which moves integrality management towards SMT solving, 
has been the only method which found a solution for the st\_miqp3 benchmark. And 
there are other cases in which the one-by-one approach may perform better than 
all-in-one (see nvs06) and cases in which binarized branch and bound performs 
better than the non-binarized one (see tln2, where the binarized all-in-one 
approach solves the problem in 0.9 seconds and the non-binarized all-in-one 
approach exceeds the timeout).

\myparagraph{Observation} Our experimental results for \Manyopt{} show clear benefits of its approach, as seen in Fig.~\ref{fig:integrality} and Fig.~\ref{fig:CRoptimization}.
If we  had delegated mixed arithmetics to the SMT solver, e.g., then we would have solved 
almost no MINLP problems instead of 68\% as there is limited support for 
the combination of mixed and non-linear arithmetics by SMT solvers. This is also 
clear from the many black squares (i.e. SMT solver returning ${\tt UNKNOWN}$) in the 
{\bf nobb} column in the table, where the mixed non-linear complexity handling 
is delegated to the SMT solving.

\subsection{Layer by layer analysis: continuous relaxation optimization}

\begin{figure}[h]
\centering
\begin{scriptsize}
\begin{tabular}{|l|r|r|r|}
\hline
\textbf{benchmark}     & \textbf{U.B.S}    & \textbf{Naive}    & \textbf{Hybrid} \\
\hline                                                                           
st\_e19                 &    0.19           &    1.02           &     0.34        \\
chance                 &    0.26           &    1.63           &     0.53        \\
linear                 &    8.24           &    4.67           &     4.91        \\
circle                 &    3.47           &    4.92           &     4.82        \\
pointpack02            &    0.19           &  206.98           &     0.47        \\
house                  &    1.03           &    \NA            &     2.04        \\
haverly                &    0.61           &    \NA            &     1.24        \\
wastewater02m1         &    \NA            &  308.82           &     \NA         \\
dispatch               &    \NA            &   38.66           &     \NA         \\
st\_e22                 &    0.15           &    1.48           &     0.27        \\
sambal                 &  219.34           &    \NA            &   527.86        \\
immun                  &   24.17           &    \NA            &    47.19        \\
\hline                                                                           
\end{tabular}
\end{scriptsize}
\caption{Typical experimental data for three continuous relaxation
  optimization methods within \Manyopt{} on selected benchmarks:
  running times are shown in seconds and their absence indicates a 10
  minute timeout \label{fig:CRoptimization}} \end{figure}

Figure \ref{fig:CRoptimization} shows the performance of the three
features available in the Continuous Relaxation Optimization layer, namely
the unbounded binary search (U.B.S column), the naive method and our hybrid
method.

Since the Continuous Relaxation Optimization features can also be used in NLP
problems, here we show the results of some tests with NLP benchmarks, where
the timeout is 10 minutes and only three features vectors have been tested
for each benchmark.

The benchmarks in this table have been chosen to show two important
facts: the first one is the complementarity between the Unbounded Binary
Search and the Naive method. In particular, while the naive search technique,
has been shown to often work well with SAT-based optimization (see MiniSAT+),
the unbounded binary search technique is faster in scanning big intervals
containing many valid objective values. The table shows how big the gaps
between the two techniques can be. For example, one of the two
methods takes just a few seconds, while the other one even exceeds the 10 minutes
timeout (see the house, haverly, wastewater02m1 and dispatch benchmarks
for example).

The second important fact shown in the table is the benefit of the hybrid
feature in mitigating the gap between unbounded binary search and naive search,
as the hybrid feature combines them by exploiting the common invariants of their
algorithms. As the table shows, its performance is almost always (but with
some exceptions) closer to the best feature between U.B.S and Naive, and
according to the overall experiments for \Manyopt it also wins very often
(in about 20\% of cases, see the logs in the \Manyopt tool webpage 
\cite{manyoptURL} for details).

\section{Related work}
\label{section:related}
We already discussed related work in the area of SMT solving and will
therefore focus here on related work in optimization.
Optimization problems with logical formulae frequently arise in
practice; examples in process systems engineering include: choosing
between multiple possible treatment technologies in a waste water
treatment plant \cite{karuppiah:2006} and designing distillation
column configurations \cite{caballero-grossmann:2001}. Typically used
techniques in process engineering will first translate logical rules
into mathematical constraints (e.g., as detailed in
\cite{floudas1995nonlinear}) and then solve the resulting MINLP
problem; this process of formulating a mathematical optimization
problem based on logical constraints and then transforming the
resulting problem into a MINLP problem is known as Generalized
Disjunctive Programming (GDP) \cite{lee-grossmann:2001}. Our proposed
method of solving MINLP using SMT technology is therefore
complementary to GDP; GDP is an excellent formulation technology, but
transforming to a MINLP problem and then using traditional optimization methods eliminates the possibility of directly exploiting logical constraints in the original formulation.

With respect to MINLP solvers based on traditional branch-and-bound approaches 
\cite{scip,belotti-etal:2009,lin-schrage:2009,antigone,sahinidis:1996}, the 
advantage of the overall approach taken in tool \Manyopt\  is that~--~as in most 
SMT-based approaches~--~there is support for incremental strategies such as 
push/pop. Typical MINLP solvers will fathom, i.e.\ discard, regions of the tree 
as soon as they determine that this region cannot include the global solution.

\section{Conclusions and Future Work}
\label{section:conclusion}

We have presented \Manyopt, an extensible tool capable of solving non-linear and 
mixed optimization problems by relying on SMT solving, thanks to the parallel
execution of complementary reduction techniques. While some of those reductions
are inspired by existing techniques, other reductions we developed and
integrated into \Manyopt{}~--~such as binarized flattening~--~are
novel and crucial for leveraging the optimization
capabilities of our tool. The experimental results confirm the effectiveness
of our tool in solving MINLP problems, which are not solvable using other
state-of-the-art SMT-based optimizers, as these have almost no support for mixed
non-linear arithmetic.

The tool's architecture is presented in layers that contain a number
of features, and the tool is extensible by adding more features into
such layers for further experimentation. In future work, we mean to
extend \Manyopt{} in this manner and test these new features and their
combinations with further experiments, for example by including the
SMT optimizer \ZTopt\ in the continuous relaxation optimization
layer. This is interesting as this would not require \ZTopt\ to deal
with mixed arithmetic but only with non-linear arithmetic.

We also would like to implement the capability of \emph{dynamic} parallel 
execution in which, instead of statically determining feature vectors a priori, 
the fastest features are chosen dynamically during the execution. For example, 
while executing the all-in-one method, the program may run the naive, unbounded 
binary search and hybrid methods, and it may then obtain the solution by a different 
method in each all-in-one iteration.

Another important line of investigation is to understand how features
used in parallel executions can share information to use such learned
insights for improving the
precision and performance of \Manyopt{} executions.

Our tool can easily incorporate external constraints to a MINLP problem, as such problems are represented in SMTLIB 2.0. We will collaborate with Chemical Process Engineers to explore the benefits of such enrichments of MINLP problems, in terms of decreasing running times, decreasing the frequency of ${\tt UNKNOWN}$ replies, and enriching the problem specification so that ``what-if'' questions can be answered with supporting witness information. 

We would like to exploit the POSIX piping within \Manyopt{}
to support other external SMT solvers as feasibility checkers, notably
some SMT solvers specifically developed for some subclasses of non-linear real
arithmetics \cite{nonlinear3,nonlinear2,nonlinear1}. Although such
solvers are currently designed for real arithmetics only, they can be used
in our tool, since we would use them to check the feasibility of continuous
relaxations of MINLP problems.

Finally, we would like to understand how search methods from
Artificial Intelligence, for example Conflict-directed A*
\cite{williams07}, could interact with~--~or be integrated in~--~our
tool and the approaches presented in this paper.

\medskip
{\bf Open-access Code and Research Data:} The code for tool \Manyopt{}
and more information on our experimental data and results are available at

\medskip
\begin{center}
${\tt bitbucket.org/andreacalliadiddio/manyopt}$
\end{center}

\medskip
{\bf Acknowledgments:} This work was supported by the UK
EPSRC with Fees Award and grants EP/N023242/1 and EP/N020030/1. We
thank Ruth Misener and Miten Mistry for discussions and advice on MINLP and the
use of SMT in solving such problems. Suleeporn Sujichantararat wrote a
MSc thesis \cite{suleeporn15}, supervised by the second author, that
shadowed some aspects of the research program for the tool reported in this paper.

\end{document}